\pdfoutput=1
\documentclass[11pt]{article}

\usepackage[preprint]{acl}

\usepackage{times}
\usepackage{latexsym}
\usepackage{siunitx}
\usepackage{amsmath}
\usepackage[T1]{fontenc}
\usepackage[utf8]{inputenc}

\usepackage{microtype}

\usepackage{inconsolata}
\usepackage{multirow}
\usepackage{framed}
\usepackage{makecell} 
\usepackage{lipsum} 
\usepackage{enumitem}
\usepackage{listings}
\usepackage{paralist}
\usepackage{booktabs}
\usepackage{graphicx}
\usepackage{amssymb}
\usepackage{pifont}
\usepackage{marginnote}
\usepackage[color=white,size=footnotesize]{todonotes}
\usepackage{verbatim}

\title{Identifying and Answering Questions with False Assumptions:\\An Interpretable Approach}

 \author{Zijie Wang \and Eduardo Blanco \\
        Department of Computer Science\\
         University of Arizona \\ 
         \texttt{\{zijiewang,eduardoblanco\}@arizona.edu}}

\begin{document}
\maketitle
\begin{abstract}
People often ask questions with false assumptions, a type of question that does not have regular answers. 
Answering such questions requires first identifying the false assumptions. 
Large Language Models (LLMs) often generate misleading answers to these questions because of hallucinations. 
In this paper, we focus on identifying and answering questions with false assumptions in several domains.
We first investigate whether the problem reduces to fact verification.
Then, we present an approach leveraging external evidence to mitigate hallucinations. 
Experiments with five LLMs demonstrate that
(1)~incorporating retrieved evidence is beneficial
and 
(2)~generating and validating atomic assumptions yields more improvements 
and provides an interpretable answer by pinpointing the false assumptions.

\end{abstract}

\section{Introduction}
Large Language Models (LLMs)~\cite{brown2020language, team2023gemini}
have demonstrated remarkable abilities in
extractive~\cite{rajpurkar-etal-2016-squad, kwiatkowski-etal-2019-natural} and generative question answering~\cite{reddy-etal-2019-coqa, fan-etal-2019-eli5} among others.
Despite their capabilities, LLMs suffer from hallucinations~\cite{zhang2023siren}.
This leads to unfaithful answers due to overconfidence---when LLMs lack knowledge, they often make up answers
despite abstaining (e.g., ``I do not know'') being more desirable~\cite{feng-etal-2024-dont}.
Overconfidence also leads to hallucinated answers to unanswerable questions~\cite{slobodkin-etal-2023-curious}.
For example, ``Which countries border Kansas?'' should be addressed by pointing out that Kansas does not have an international border.

\begin{figure}
  \centering
  \small
  \includegraphics[width=\columnwidth]{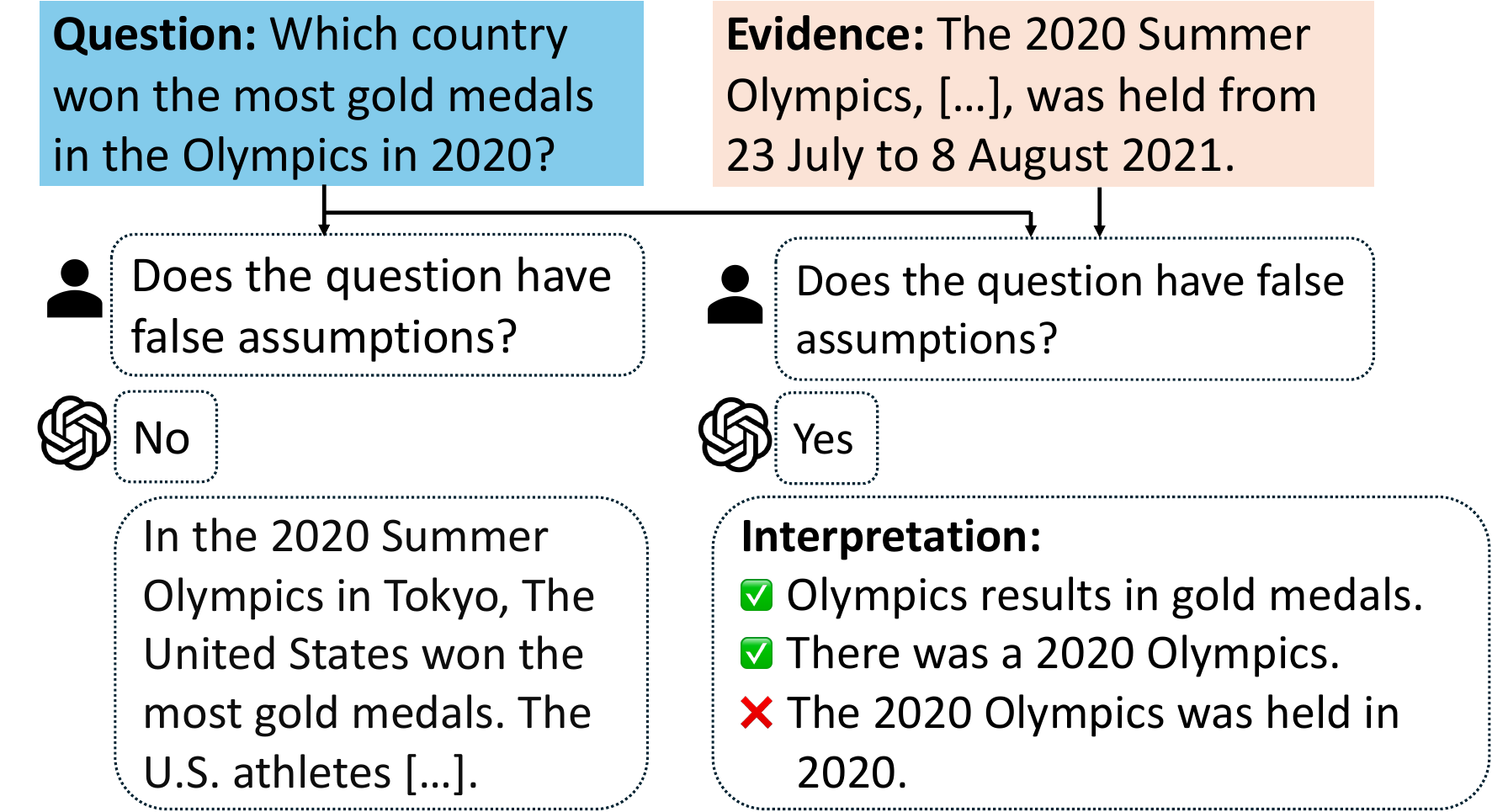}
  \caption{
    Question with a false assumption (i.e., the 2020 Olympics was held in 2020).
    ChatGPT provides a misleading answer without identifying the false assumption.
    Our approach couples external evidence with a process to generate and validate atomic assumptions.
    Doing so allows us to answer questions with false assumptions by pointing out the false assumptions
    (e.g., There was a 2020 Olympics but it was not held in 2020).}
    \label{f:motivation}
\end{figure}

Unanswerable questions, as studied in previous works~\cite{kim-etal-2023-qa,yu-etal-2023-crepe,hu-etal-2023-wont},
are information-seeking questions containing false assumptions.  
The false assumptions make them lack regular answers.
Instead, answers should point out the false assumptions.
Consider the question in Figure~\ref{f:motivation}.
It wrongly assumes that the 2020 Olympics was held in 2020.
The event, however, was held in 2021.  
ChatGPT fails to identify the false assumption and generates a misleading answer.
In this paper, we investigate the problem of identifying and answering questions with false assumptions.
We define false assumptions in questions as subjective opinions, beliefs, or misconceptions 
held by the question author. Note that they differ from false facts, which are objective claims 
contradicting reality. 

By leveraging retrieval-augmented methods, our approach (Figure~\ref{f:motivation}, right) mitigates 
hallucinations.
In contrast to previous works aiming to generate free-form answers to 
questions with false assumptions~\cite{kim-etal-2023-qa,yu-etal-2023-crepe,hu-etal-2023-wont},  
we propose to generate and validate atomic assumptions. 
Doing so not only benefits identifying false assumptions but also pinpoints the false assumptions---often a small part of the question---thus yielding a human-interpretable answer.

As the examples above and Table~\ref{t:dataset_example} show,
identifying and answering questions with false assumptions is challenging.
Our main contributions are:\footnote{Code and dataset available at \url{https://github.com/wang-zijie/Question-with-False-Assumption}}
\begin{compactitem}
  \item A set of general-purpose approaches to identify questions with false assumptions. 
  \item Interpretable answers derived from the generation and validation of atomic assumptions.
  \item Experiments showing that our approach yields state-of-the-art results across three datasets yet requires less compute than other approaches.
  \item Error analysis providing insights into the questions and false assumptions that lead to misclassifications by our best model.
\end{compactitem}

\section{Related Work and Existing Datasets}
\label{s:related_work}

\paragraph{Answering Special Questions}
Question answering has evolved from answering factoid questions~\cite{clark-etal-2019-boolq} to non-factoid questions~\cite{soleimani-etal-2021-nlquad};
from reading comprehension to open-domain QA~\cite{karpukhin-etal-2020-dense}.
Recently, research efforts have been made on various special types of questions.
% \citet{chen2021dataset} develop a time-sensitive dataset including questions that expect different answers at different times. For example, 
% ``When will season 2 of House of the Dragon be released?'' can only be answered after 2024. 
For example, \citet{min-etal-2020-ambigqa}
investigate ambiguous questions (i.e., questions with more than one valid answer).
\citet{stelmakh-etal-2022-asqa} further extend it by answering the questions with long-form responses.
\citet{wang-etal-2023-interpreting,wang-etal-2024-interpreting} interpret answers to yes-no questions that do not contain \textit{yes} or \textit{no}.
% \citet{yu-etal-2023-ifqa} modify questions with \emph{if} counterfactuals;
% original answers sometimes---not always---must be updated.

\paragraph{Retrieval-Augmented LLMs}
Document retrieval has been used in question answering for decades~\cite{moldovan-etal-2002-performance}.
Dense Passage Retriever~\cite{karpukhin-etal-2020-dense} leverages retrieval and LLMs.
Retrieval-Augmented Generation~\cite[RAG]{lewis2020retrieval,guu2020retrieval} combines retrieval and generation models
% to generate outputs,
to mitigate hallucinations%
%thereby mitigating hallucinations with the aid of external knowledge~%
%Such method is capable of mitigating LLMs hallucinations by leveraging external knowledge
~\cite{gao2023retrieval}. 
In addition, retrieval-augmented methods have been integrated into LLMs. % in different ways.
% \citet{zhang2024raft} propose a LLM fine-tuning method augmenting retrieved documents.
\citet{chen2022decoupling}
incorporate retrieved instances from the training corpus into prompts.
\citet{peng-etal-2025-unanswerability} evaluate RAG systems on scenarios in which the queries are unanswerable based on the given knowledge base.
Unlike previous works, we focus on identifying and answering questions with false assumptions.
%We demostrate thatand our approach demonstrates that incorporating retrieved evidence is beneficial.

\paragraph{Fact Verification}
Verifying facts is typically framed as the problem of determining whether a claim is supported or contradicted by a source document~\cite{chen2023complex}.
Fact verification and identifying questions with false assumptions are only distantly related,
as questions with false assumptions often do not have wrong facts.
For example, ``Why can bald people grow beards?'' does not challenge the veracity of ``Bald people can grow beards.''
The question, however, has a false assumption: head hair is the same as facial hair.
As we shall see,
identifying false assumptions cannot be reduced to fact verification (Section \ref{s:experiments}).

\paragraph{Existing Datasets} %\todo{merge related work with existing datasets, remove Terminology to save space}
Several works present datasets consisting of questions with false assumptions: 
(QA)$^2$~\cite{kim-etal-2023-qa}, CREPE~\cite{yu-etal-2023-crepe}, and FalseQA~\cite{hu-etal-2023-wont}. 
(QA)$^2$ obtains questions from Google %\footnote{\url{https://developers.google.com/maps/documentation/places/web-service/query}} 
and asks crowdworkers to
(1)~identify whether the questions have false assumptions
and
(2)~write answers. 
CREPE obtains questions and answers from the ELI5 subreddit dataset~\cite{fan-etal-2019-eli5}.
%They extract the highest-vote replies in the thread as the gold answers to the questions. 
Like (QA)$^2$, they rely on crowdworkers to identify false assumptions in the questions.
FalseQA consists of synthetic questions about several topics written by crowdworkers, who are instructed to make up false assumptions.
%They are asked to write questions with false assumptions covering several topics and then revise
%them into questions with valid assumptions.
This results in questions with few variations. Answers to questions are also written manually.

\begin{table}
  \centering
  \small
  \begin{tabular}{@{}l rrr@{}}

\toprule

& (QA)$^2$ & CREPE & FalseQA \\
\midrule
Genuine questions?             & \ding{51}  & \ding{51}  & \ding{55} \\
Genuine answers?            & \ding{55}  & \ding{51}  & \ding{55} \\
Human-written evidence?     & \ding{51}  & \ding{51}  & \ding{55}\\
Auto-retrieved evidence?    & \ding{55}  & \ding{51}  & \ding{55}\\ \addlinespace

\# instances                &  602  & 8,444 & 4,730 \\
~~~~\# train                &   32  & 3,462 & 2,374 \\
~~~~\# validation           &  n/a  & 2,000 &  982  \\
~~~~\# test                 &  570  & 3,004 & 1,374 \\
~~~~~~\% valid assumptions  &   50  &   75  & 50    \\
~~~~~~\% false assumptions  &   50  &   25  & 50    \\

\bottomrule

\end{tabular}
  \caption{
    Existing corpora containing questions with false assumptions.
    %FalseQA does not source questions from genuine texts---annotators wrote questions on demand---and CREPE is the only one with genuine answers.
    These corpora target different domains (search logs, Reddit and selected topics). % respectively).
    %and CREPE is not balanced.
  }
  \label{t:dataset}
\end{table}

Table~\ref{t:dataset} presents basic information about the three datasets. 
FalseQA is the only one without genuine questions (i.e., annotator-written on demand) and provides no external evidence. 
In contrast, (QA)$^2$ and CREPE contain genuine questions and include evidence written by crowdworkers. 
CREPE further provides passages retrieved by C-REALM~\cite{krishna-etal-2021-hurdles}.
% a retrieval model trained on Wikipedia corpora.
Notably, (QA)$^2$ has relatively few instances compared to the other two datasets and lacks training instances.
CREPE is the only corpus that
(1)~contains genuine answers
and
(2)~has an unbalanced label distribution.

\begin{figure*}[htp!]
  \centering
  \small
  \includegraphics[width=1\textwidth]{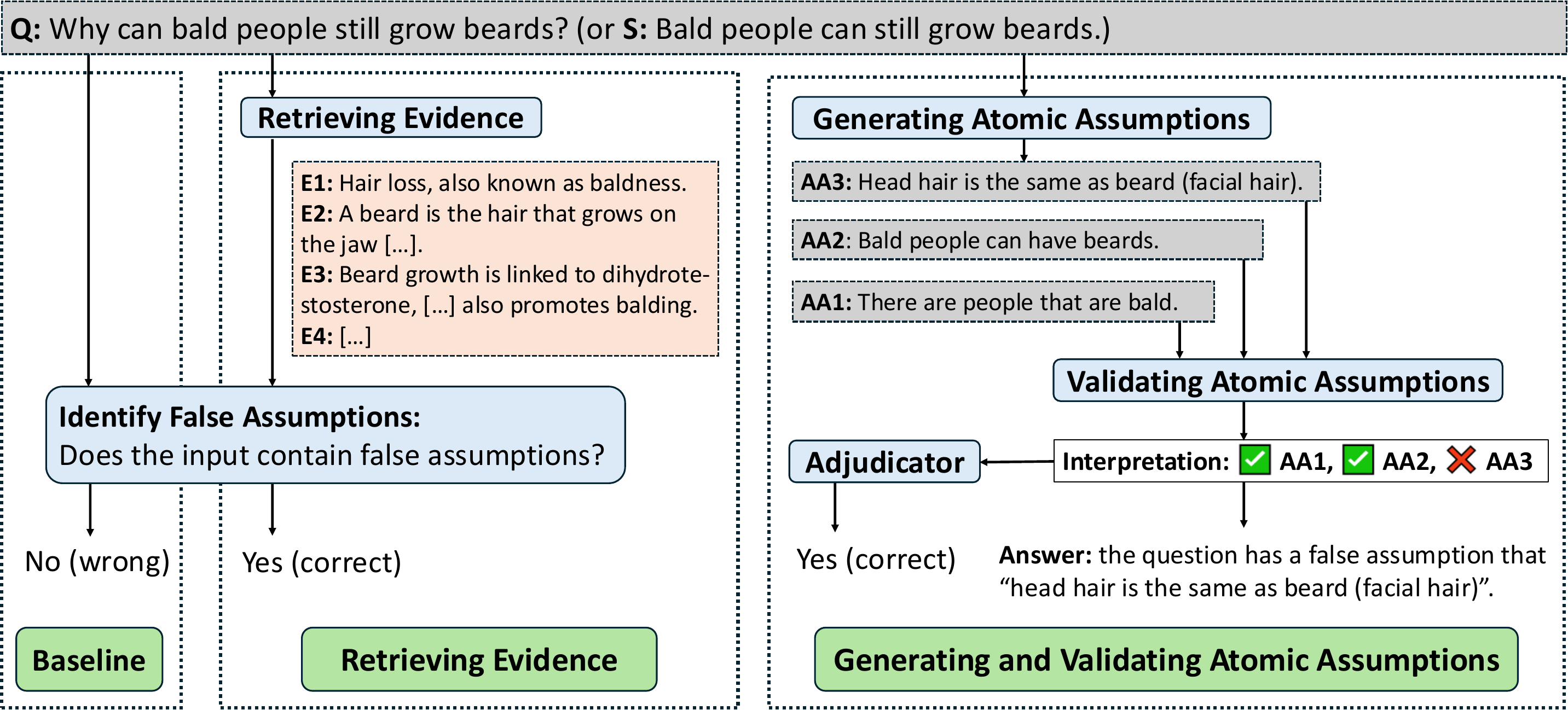}
  \caption{Approaches to identify and answer questions with false assumptions.
    Baselines only have access to the question or statement.
    Approaches retrieving evidence incorporate relevant information. 
    Generating and validating atomic assumptions
    yields human-readable interpretations as well as answers to the question. %(i.e., the atomic assumption that is false, if any).
    }
    
    \label{f:pipeline}
\end{figure*}
Beyond these datasets, \citet{zhao-etal-2024-couldve} present a benchmark to evaluate LLMs' ability on rewriting the unanswerable questions from (QA)$^2$.
\citet{yang2024towards} investigate unanswerable questions in the Electronic Health Records domain.

Despite previous efforts on this problem,  
we are the first to propose a unified approach that identifies and answers questions with false assumptions across all existing datasets.
More importantly, previous works neither specify the false assumptions nor provide any interpretations or insights to correct the false assumptions.
We propose to generate and validate atomic assumptions.
This approach allows us to
(1) provide interpretations for the binary identification task
and 
(2) answer questions with false assumptions by pinpointing specific false assumptions.

\section{Methods to Identify and Answer Questions with False Assumptions}
\label{s:method}

We define the task of identifying and answering questions with false assumptions as follows.
Consider a question $Q$, its reformation as a statement $S$, 
a set of atomic assumptions $AA = [AA_1, \ldots, AA_n]$ generated from $Q$, 
a set of evidence $E$ retrieved based on $Q$ or $S$, 
and a label set $L \in [0,1]$,  where: 
% \vspace{-0.5em}
\begin{equation*}
L = \begin{cases}
     0  & \text{(contains false assumptions)} \\
     1  & \text{(contains no false assumptions)}
  \end{cases}
\end{equation*}

Identifying questions with false assumptions is to
learn a mapping $f: (I, E) \mapsto L$, with $I \in [Q, S, AA]$ and $E$ might be empty.  
We answer the questions by first validating the set of atomic assumptions $L_{AA} = [AA_1: L_1, \ldots, AA_n: L_n]$, and then generating an interpretable answer $A$ by verbalizing $L_{AA}$.

We present four baselines: (1) reducing the problem of identifying false assumptions to fact verification, 
(2) supervised fine-tuning language models, 
(3) prompting LLMs, 
and
(4) incorporating generated evidence into prompting~\cite{liu-etal-2022-generated} (Figure~\ref{f:pipeline}, left block).
Then, we present methods to
(1)~consider retrieved evidence (Figure~\ref{f:pipeline}, middle block)
and
(2)~generate and validate atomic assumptions (Figure~\ref{f:pipeline}, right block).

\subsection{Baselines}
 \label{ss:baseline}

\paragraph{Reducing the Problem to Fact Verification}
Fact verification aims at verifying facts grounded on support documents~\cite{guo-etal-2022-survey}.
At first sight, identifying false assumptions could be solved by
transforming questions into statements
and
finding falsehoods in the corresponding statements.

\begin{table*}
  \centering
  \small
  \begin{tabular}{@{}p{2.98in}p{3.1in}@{}}

\toprule

\textit{Q1}: Why are ice cubes mostly clear but icebergs are white? 
& \textit{Statement}: Ice cubes are mostly clear and icebergs are white. \\ \addlinespace

\textit{Evidence}: Commercially made ice cubes may be clear; Icebergs are generally white because they are covered in snow; 
  Although ice by itself is clear, snow usually appears white in color due to diffuse reflection [\ldots]. 
& \textit{Atomic Assumptions}:
  (a1 \ding{51}) Ice cubes and icebergs are made of water. %the same material (water).
  (a2 \ding{51}) Ice cubes are mostly clear.
  (a3 \ding{51}) Icebergs are white.
  (a4 \ding{51}) Ice cubes and icebergs can be different in color despite they are made of the same material. \\ \midrule

\textit{Q2}: When did the San Andreas Fault last erupt?
& \textit{Statement}: The San Andreas Fault has erupted before. \\ \addlinespace

\textit{Evidence}: The San Andreas Fault is a transform fault; 
  Transform fault involves no loss of lithosphere at [\ldots]; 
  Most volcanic activity happens where lithosphere is being destroyed.
& \textit{Atomic Assumptions}:
  %(a1 \ding{51}) There is a geological feature calledthe San Andreas Fault.
  (a1 \ding{51}) The San Andreas Fault is a geological feature.
%  (2) The San Andreas Fault is capable of erupting.
  (a2 \ding{55}) The San Andreas Fault can erupt. %is capable of erupting.
  (a3 \ding{55}) The San Andreas Fault has erupted during a known time. \\ \midrule

\textit{Q3}: When did they stop using lead in pencils?
& \textit{Statement}: People stopped using lead in pencils. \\ \addlinespace

\textit{Evidence}: [\ldots] lead has not been used for writing [\ldots];
  Because the pencil core is still referred to as ``lead'',
  people have the misconception that the graphite in the pencil is lead.
& \textit{Atomic Assumptions}:
  (a1 \ding{55}) Pencils were once made using lead.
  (a2 \ding{51}) Pencils no longer contain lead. 
  (a3 \ding{55}) There was a specific time when people stopped using lead in pencils. \\ \bottomrule  
\end{tabular}

  \caption{Three questions with false assumptions from (QA)$^2$ and CREPE.
    Our approach automatically transforms the questions into (1) a single statement and (2) generates and validates atomic assumptions (\ding{51} or \ding{55}).
    Further, we retrieve evidence from external sources as doing so is more beneficial than generating relevant evidence with LLMs.
    }
  \label{t:dataset_example}

\end{table*}

As described next, prompting LLMs is successful at transforming questions into statements.
Verifying the statements, however, does not equate to identifying false assumptions in the questions.
This is because they are fundamentally different problems.
As we discussed before, identifying false assumptions is different from verifying false facts.
For example,
\emph{How can we see the moon in the middle of the day?}
does not challenge the truth of ``The moon is visible in the middle of the day''. In fact, the question indicates that the author is aware that the statement is true.
The question, however, also falsely assumes that ``The moon is expected to be seen only at night''.
As we shall see, these two problems are only distantly related (Section \ref{s:experiments}).

\paragraph{Transforming Questions into Statements}
A few-shot prompt with GPT-4 is sufficient to reliably transform question $Q$ into statement $S$. 
Appendix \ref{ss:appendixb} lists the full prompts.
We evaluated the transformation quality by manually checking 200 question-statement pairs from each corpus (600 total).
The correctness of the transformation is high for all datasets
((QA)$^2$: 0.94, CREPE: 0.89, and FalseQA: 0.98).
Table~\ref{t:dataset_example} shows three more examples of questions transformed into statements.

\paragraph{Supervised Approach}
As discussed in Section~\ref{s:related_work}, CREPE and FalseQA provide training splits. 
(QA)$^2$ only includes 32 instances for in-context learning. 
We first investigate a cross-domain transfer learning baseline by training a model using CREPE and (or) FalseQA and evaluate with the three datasets: (QA)$^2$, CREPE, and FalseQA.
Additionally, we explore another two relevant datasets: BoolQ~\cite{clark-etal-2019-boolq} and FEVER~\cite{thorne-etal-2018-fever}. 
Since \citet{yu-etal-2023-crepe} demonstrate that MNLI~\cite{williams-etal-2018-broad} yields worse results, we do not include any NLI datasets.
% \begin{figure}
%   \small
%   \begin{framed}
%   \texttt{You will be provided with an input containing a \{question or statement\}. Your task is to identify if there are false assumptions in the input.
%   Answer Yes if the input contains any false assumptions. Otherwise, answer No.}\\\\
%   \texttt{\{4-shot examples\}}\\\\
%   \texttt{Input: \textit{\{question or statement\}}}\\
%   \texttt{Does the Input contain any false assumptions?}

%   \end{framed}
%   \caption{Prompts to identify questions or statements.}
%       \label{f:prompts}
%   \end{figure}

\paragraph{Prompting LLMs}
The second baseline involves prompting LLMs to identify if a question has false assumptions. 
These prompts only rely on LLMs' internal knowledge acquired during pretraining and are affected by hallucinations.
We use a few-shot prompt asking whether the question (or statement automatically generated, Section \ref{ss:baseline}) has false assumptions. 
%On the other hand, we also prompt to identify if the statement generated for fact verification (Section~\ref{ss:fact_check_baseline}) has false assumption, 
% As we shall see, such transformation loses essential information and thus make the identification impossible.
Appendix~\ref{ss:appendixa} provides the complete prompts.

% \paragraph{Generated Knowledge Prompting}

\paragraph{Prompting LLMs with Generated Evidence}
%Incorporating evidence is beneficial for answering questions~\cite{mitra2019additional}.
Following \citet{liu-etal-2022-generated},
we
(1) generate evidence from the question or statement using an LLM
and 
(2) incorporate the generated evidence into the prompt.
%We adopt a similar approach that generates evidence with LLMs
%from the question or statement,
%and then incorporate the evidence into the prompts.
Note that the generated evidence is likely to include hallucinations.
This baseline allows us to determine whether retrieving external evidence
outperforms evidence generated by the LLM itself.
As we shall see, generated evidence is detrimental for this task while retrieved evidence is beneficial.
%retrieved evidence is our approach that retrieves external evidence outperforms generated evidence which is bounded by LLMs' internal knowledge. (It does).
%We reuse the prompts by~\citet{liu-etal-2022-generated} with minimal modifications to generate evidence.
Detailed prompts can be found in Appendix~\ref{ss:appendixc}.

\paragraph{Leveraging LLMs with Complex Reasoning Ability}
State-of-the-art LLMs such as OpenAI o1 have shown to be capable of complex reasoning tasks~\cite{jaech2024openai}.
We prompt the o1 model to identify questions with false assumptions using similar but zero-shot prompting as the previous baselines.
This practice is suggested by the model author to provide simple yet clear instructions.

\subsection{Retrieving Evidence}
\label{ss:retrieving}

Retrieval-augmented methods~\cite{lewis2020retrieval,guu2020retrieval} mitigate hallucinations by obtaining knowledge from external sources---up-to-date knowledge can be readily retrieved.
As shown in Figure~\ref{f:pipeline} (middle block), we propose a retrieval-augmented method to identify questions with false assumptions by
(1) retrieving an evidence set $E$ based on $Q$ or $S$
and
(2) incorporating the evidence in the prompts.

% Since retrieval-augmented methods have been widely used in many NLP tasks, we skip the details to retrieve the evidence here and
% report them in. 

\paragraph{Retrieving Documents}
We begin by retrieving documents relevant to the question or statement.
For CREPE, we reuse the relevant Wikipedia passages retrieved with C-REALM~\cite{krishna-etal-2021-hurdles}.
For (QA)$^2$ and FalseQA, we query the Google Search Engine API,
restricting results to Wikipedia articles and retaining the top three.

\paragraph{Retrieving Sentences}
From the retrieved Wikipedia articles,
we consider all sentences as candidate evidence.
We employ INSTRUCTOR~\cite{su-etal-2023-one}, a state-of-the-art text embedding model, to identify the top $k$ sentences most relevant to the input question or statement.
We treat $k$ as a tunable parameter (maximum: 10).
The quality of retrieved evidence for identifying false assumptions is evaluated empirically through our experimental results.
Table~\ref{t:dataset_example} illustrates questions with their corresponding retrieved evidence.
For instance, evidence for ``When did they stop using lead in pencils?''
includes ``[\ldots] lead has not been used for writing'' which directly contradicts the false assumption in the question (i.e., people once used lead in pencils).
The experimental settings are reported in Appendix~\ref{ss:appendixe}, and the complete prompts incorporating retrieved evidence are provided in Appendix~\ref{ss:appendixd}. 

\subsection{Generating and Validating Atomic Assumptions}
\label{ss:generating}

The identification methods discussed so far indicate whether a question contains false assumptions---$L \in \{0,1\}$.
They neither specify the false assumptions nor provide any interpretations or insights to correct the false assumptions.

As shown in Figure~\ref{f:pipeline} (right block),
our method grounded on generating and validating atomic assumptions pinpoints specific false (and true) assumptions in questions.
Atomic assumptions are explicit and implicit elemental information
stated and believed to be true by the question author,
and provide human-readable interpretations for the identification problem.
Further, we argue that validated atomic assumptions (true or false) are more sound answers to questions with false assumptions than the free-form answers generated in previous work.
This is because comparing a single gold answer to the generated one with existing metrics such as
BLEU~\cite{papineni-etal-2002-bleu} cannot effectively assess correctness~\cite{min-etal-2023-factscore}.
Further, they cannot distinguish when answers generate plausible answers without realizing the question has false assumptions.
For example, ChatGPT's answer to the question from Figure~\ref{f:motivation} 
(``United States won the most gold medals in the 2020 Olympics'') is factually correct but does not address the false assumption:
\textit{The 2020 Olympics was held in 2020}. 
In fact, it was held in 2021.
On the other hand, generating and validating atomic assumptions does address the false assumptions in the question: There was a 2020 Olympics but it was not held in 2020.
Table \ref{t:dataset_example} and Figure \ref{f:pipeline} detail more examples.

We generate a set of atomic assumptions $AA$ from $Q$ 
and then validate each $AA_i \in AA$ to obtain $L_{AA}$. 
The last step is to aggregate $L_{AA}$ to determine whether
the question has a false assumption, in another word, $L$.
We propose a simple adjudicator: $L = \bigwedge_{i=1}^n L_i$, $L_i \in L_{AA}$.
As we shall see, this approach (1) yields improvements in identifying false assumptions, 
(2) provides interpretations for the identification task,
and (3) is more efficient than other methods to obtain the interpretation.

It is more beneficial to generate atomic assumptions from the question than the statement by
leveraging GPT-4o with a few-shot chain-of-thought prompt~\cite{wei2022chain}. 
Note that we first experiment with GPT-4o to generate the atomic assumptions. Later we present results with smaller LLMs.  
This task is more challenging than similar practice that extracts atomic facts or claims from 
long-form text generation~\cite{min-etal-2023-factscore} or book-length summarization~\cite{fables-2024-kim-et-al}, 
as it requires extracting both explicit and implicit information behind a relatively short question.

Appendix~\ref{ss:appendixf} reports the specific prompts and results of a quality check process.
Overall, the precision of generated atomic assumptions is near perfect for all three datasets. 
Note that we are unable to calculate the recall as it is unclear what the full list of atomic assumptions is (e.g., is it worth generating \emph{Water can freeze in the shape of a cube} from Q1 in Table \ref{t:dataset_example}?).
On average, we generate~4.7 atomic assumptions per question.

Validating atomic assumptions is conceptually the same as validating the statements derived from questions.
We reuse the same prompts described in Section \ref{ss:baseline} and
Section \ref{ss:retrieving}.
\section{Experiments and Results}
\label{s:experiments}

We are the first to experiment across all three corpora including questions with false assumptions:
(QA)$^2$, CREPE, and FalseQA.
As we shall see,
generating and validating atomic assumptions to identify and answer questions with false assumptions
(1)~is the best performing across the three corpora
and
(2)~unlike previous work, it is interpretable by design.
The first row in Table \ref{t:baseline} presents the best results to date with each corpus.
Note that these systems are crafted for each corpus;
unlike us, the authors do not conduct any cross-corpora evaluation.

\begin{table}[t!]
  \centering
  \small
  \setlength{\tabcolsep}{.04in}

\begin{tabular}{l rrr}

\toprule
 & (QA)$^2$ & CREPE & FalseQA \\
& (Acc)    & (F1)  & (Acc)   \\ \midrule

Best Prev. Work & 64.00  & 67.00  & 86.00  \\ 
            & \shortcite{kim-etal-2023-qa} & \shortcite{yu-etal-2023-crepe} & \shortcite{hu-etal-2023-wont} \\
\midrule

Fact Verification w/ Question \\
~~~and Evidence  \\
~~~~~~Gold (upper bound)        & 72.93  & 49.47  & n/a  \\
~~~~~~Retrieved w/ Question     & 69.47  & 42.41  & 65.57\\
~~~~~~Retrieved w/ Statement    & 68.25  & 41.31  & 63.32\\\addlinespace

Fact Verification w/ Statement \\ 
~~~and Evidence  \\
~~~~~~Gold                      & 75.71  & 54.36  & n/a  \\
~~~~~~Retrieved w/ Question     & 73.33  & 52.58  & 70.89  \\
~~~~~~Retrieved w/ Statement    & 71.75  & 52.06  & 71.18\\
\midrule
Best Fine-tuned RoBERTa  & 56.07 & 62.81  & 73.59\\
% ~~~CREPE           & 0.50  & 0.60  & 0.50  \\
% ~~~FalseQA         & 0.56  & 0.55  & 0.71  \\
% ~~~CREPE + FalseQA & 0.49  & 0.62  & 0.52  \\
% ~~~~~~+ BoolQ      & 0.52  & 0.52  & 0.72  \\
% ~~~~~~+ FEVER      & 0.55  & 0.58  & 0.71  \\
% ~~~All             & 0.52  & 0.59  & 0.73  \\
 \addlinespace

Prompting GPT-4o with  \\
~~~Question  & 72.32  & 62.62  & 73.77  \\
~~~Statement & 71.05  & 59.93  & 71.90  \\ 
~~~Generated Evidence & 67.03 & 61.24 & 73.57 \\  \addlinespace
% ~~~~~~0-shot & 66.57 & 61.24 & 73.45 \\
% ~~~~~~4-shot & 67.03 & 61.24 & 73.57 \\ \addlinespace

Prompting OpenAI o1 with  \\
~~~Question  & \textbf{75.08}  & \textbf{67.24}  & \textbf{80.81}  \\
~~~Statement & 74.19  & 64.13  & 76.06  \\

\bottomrule

\end{tabular}
  \caption{
    Results obtained with
    (1) the best previous work for each corpus,
    (2) a state-of-the-art fact verification system~\cite{tang-etal-2024-minicheck},
    and
    (3) several baselines.
    Fact verification with the statement yields better results than the question,
    but it underperforms even the simplest prompting.
    RoBERTa underperforms,
    and evidence generated~\cite{liu-etal-2022-generated} with GPT-4o is detrimental.
    Prompting with o1 model yields the best results (bold). }
  \label{t:baseline}
\end{table}

\begin{table*}[htp!]
  \centering
  \small

\setlength{\tabcolsep}{.052 in}

\begin{tabular}{
    l 
    r@{}l r@{}l r@{}l
    r@{}l r@{}l r@{}l
    r@{}l r@{}l r@{}l}
    
    \toprule
    & \multicolumn{6}{c}{GPT-4o} & \multicolumn{6}{c}{Llama 3 70B} & \multicolumn{6}{c}{Mistral 7B} \\
    \cmidrule(r){2-7} \cmidrule(r){8-13} \cmidrule(r){14-19}
    & \multicolumn{2}{c}{(QA)$^2$} & \multicolumn{2}{c}{CREPE} & \multicolumn{2}{c}{FalseQA} & 
        \multicolumn{2}{c}{(QA)$^2$} & \multicolumn{2}{c}{CREPE} & \multicolumn{2}{c}{FalseQA} &
        \multicolumn{2}{c}{(QA)$^2$} & \multicolumn{2}{c}{CREPE} & \multicolumn{2}{c}{FalseQA} \\
    & \multicolumn{2}{c}{(Acc)} & \multicolumn{2}{c}{(F1)} & \multicolumn{2}{c}{(Acc)} &
        \multicolumn{2}{c}{(Acc)} & \multicolumn{2}{c}{(F1)} & \multicolumn{2}{c}{(Acc)} &
        \multicolumn{2}{c}{(Acc)} & \multicolumn{2}{c}{(F1)} & \multicolumn{2}{c}{(Acc)} \\
        % \cmidrule(r){1-1}\cmidrule(r){2-7} \cmidrule(r){8-13} \cmidrule(r){14-19}
    \midrule
    % Best Prev. Work  & 0.64  & & 0.67 & & 0.86   & & 0.64 & & 0.67 & & 0.86      & & 0.64 & & 0.67 & & 0.86 &\\ \midrule
    
    Best from Baselines        & 75.08   & & 67.24 & & 80.81   & &                     75.08 & & 67.24 & & 80.81   & &   75.08 & & 67.24 & & 80.81  & \\
    \midrule
    Identifying with Question        \\ 
    ~~~w/o Evidence (baseline)   & 72.32  & & 62.62 & & 73.77   & &                     55.37 & & 52.48 & & 72.93   & &         50.72 & & 48.57 & & 57.20 & \\ 
    ~~~with Retrieved Evidence  \\ 
    ~~~~~~Gold  (upper bound)    & 85.96  &*& 74.10 &*& {n/a}   & &                     80.53 &*& 74.75 &*& {n/a}  & &          72.40 & & 58.36 &*& {n/a} & \\
    ~~~~~~using the Question     & 76.46  &*& \textbf{69.24} &*& \textbf{83.91}&*&      63.33 &*& 62.32 &*& \textbf{77.66} &*&    52.46 & & 55.23 &*& 59.04 & \\
    ~~~~~~using the Statement    & 76.21  &*& 68.24 &*& 83.49   &*&                     62.98 &*& \textbf{62.65} &*& 76.49 &*&    52.81 & & 54.80 &*& 58.84  & \\ \addlinespace

    Identifying with Statement  \\ 
    ~~~w/o Evidence (baseline)      & 71.05  & & 59.93 & & 71.90   & &                  56.84 & & 50.45 & & 70.22   & &         51.42 & & 48.13 & & 58.95 & \\ 
    ~~~with Retrieved Evidence  \\ 
    ~~~~~~Gold (upper bound)     & 84.74  &*& 70.67  &*& {n/a}   & &                    76.84 &*& 67.40 &*& {n/a}  & &          71.98 &*& 60.32 &*& {n/a} & \\
    ~~~~~~using the Question     & \textbf{76.51}  &*& 64.35  &*& 79.40   &*&           \textbf{63.68} &*& 57.85 &*& 72.20&&    54.93 &*& \textbf{55.58} &*& 59.24 & \\
    ~~~~~~using the Statement    & 76.33  &*& 63.62  & & 79.11   &*&                    61.75 && 58.14 &*& 73.07   & &          \textbf{57.21} &*& 54.89 &*& \textbf{59.95} & \\ \midrule

    Gen. \& Val. Atomic Assumptions  \\ 
    ~~~w/o Evidence (baseline)      & 71.39  & & 69.52 & & 82.88   & &                  60.53 & & 68.42 & & 74.73   & &         57.02 & & 52.20 & & 60.08 & \\
    ~~~with Retrieved Evidence  \\ 
    ~~~~~~Gold (upper bound)       & 83.81 &*& 73.52 &*& {n/a}   & &                    78.95 &*& 76.82 &*& {n/a}  & &          64.74 &*& 67.78  &*& {n/a} & \\
    ~~~~~~using the Question & \textbf{73.86}&*& \textbf{69.91} &&\textbf{86.02}&*& \textbf{65.79} &*& \textbf{70.05} &*& 85.43 &*& \textbf{60.70} &*& \textbf{53.63}  && 63.15 &* \\
    ~~~~~~using the Statement      & 72.60 && 68.24 & & 85.10    &*&                   65.79 &*& 69.91 && \textbf{85.57} &*&    60.58 &*& 52.38  && \textbf{64.48} &* \\ 

    % Gen. \& Val. Atomic Assumptions  \\ 
    % ~~~w/o Evidence (baseline)      & 71.23  & & 58.65 & & 72.25   & &   56.11 & & 54.38 & & 69.03   & &    57.32 & & 46.87 & & 57.20 & \\
    % ~~~with Retrieved Evidence  \\ 
    % ~~~~~~Gold (upper bound)       & 80.35 &*& 69.41 &*& {n/a}   & &     77.72 &*& 67.52 &*& {n/a}  & &     69.63 &*& 57.57  &*& {n/a} & \\
    % ~~~~~~using the  Question      & 76.33 &*& 63.94 &*& 80.41    &*&    62.63 &*& 60.08 &*& 72.12   & &    \textbf{60.46} & & 54.75  &*& 60.90 &* \\
    % ~~~~~~using the Statement      & 76.46 &*& 62.18 & & 81.79    &*&    62.98 &*& 59.58 &*& 69.99   & &    58.33 & & 54.21  &*& \textbf{62.30} &* \\ 

    \bottomrule
\end{tabular}

  \caption{
    Results obtained with GPT-4o, Llama 3 70B, and Mistral 7B
    (1)~prompting using the question or statement without and with evidence (middle block; without is equivalent to \textit{Prompting} in Table \ref{t:baseline})
    and
    (2)~generating and validating assumptions with and without evidence (bottom block).
    %; note that the original question and statement are disregarded).
    Including retrieved evidence is always beneficial---most improvements are statistically significant (indicated with an asterisk; McNemar's test \cite{mcnemar1947note}, p<0.05).
    Generating and validating atomic assumptions yields competitive results and, crucially,
    (1) succinct interpretations for the identification task
    and
    (2) answer to the question pinpointing the false (and true) assumptions.
%  Experiment results on verifying questions with false assumptions. We report the same metric as the original paper
%  for each dataset.
  }
  \label{t:experiment_results}
\end{table*}

\subsection{Results with Baselines}

\paragraph{Fact Verification}
We evaluate MiniCheck~\cite{tang-etal-2024-minicheck},
a state-of-the-art fact verification system,\footnote{\url{https://llm-aggrefact.github.io/}}
to identify questions with false assumptions as a fact verification task.
(QA)$^2$ and CREPE provide gold evidence that is written or retrieved by humans (Section~\ref{s:related_work});
we use it to define an (unrealistic) upper bound.
We retrieve evidence (Section \ref{ss:retrieving}) using the question or statement,
and keep the top-10 evidence sentences.

Fact verification with the statement yields better results than the question (Table \ref{t:baseline}, second block).
While it obtains somewhat high results, as we shall see, simple supervised models and prompts outperform fact verification---even with gold evidence.
We conclude that fact verification helps identifying questions with false assumptions
but the latter cannot be reduced to the former.

\paragraph{Supervised Approach}
The supervised approach finetunes a RoBERTa-large model~\cite{liu2019roberta}. 
Table~\ref{t:baseline} only reports the best results on each corpus;
Table~\ref{t:supervised_experiments} in Appendix~\ref{s:appendixb} lists the complete results.
A supervised RoBERTa model outperforms fact verification. 
However, cross-domain learning yields no improvements with CREPE and FalseQA.
In fact, when in-domain training instances are unavailable (e.g., (QA)$^2$), fine-tuning with any corpora yields similar results,   
demonstrating that the effectiveness of the supervised approach is bounded by the availability of in-domain instances.
% We note that there is no training instances for (QA)$^2$, and training with other corpora is barely beneficial.

\paragraph{Prompting without Retrieved Evidence}
Prompting GPT-4o
%\footnote{\url{https://openai.com/index/gpt-4o-system-card/}}
without evidence outperforms both fact verification and the supervised model,
although the benefits are minimal if training data is available (CREPE, FalseQA).
Thus, simple prompting is not justified if training data is available,
as a small, finetuned RoBERTa obtains virtually the same results.
Contrary to previous work~\cite{liu-etal-2022-generated},
we observe that incorporating generated evidence (not retrieved) with GPT-4o is detrimental for our task.
The drops are substantial with (QA)$^2$: 67.03 vs. 72.32.
We hypothesize that this is due to LLMs often reciting information provided to them
even if it is incorrect~\cite{wu2024faithful}. 
Prompting with the more powerful o1 model yields the best results, which outperforms the best results from previous works for two datasets.
However, as we shall see, this approach requires substantial computational costs and still falls behind our approach.

\subsection{Results Retrieving Evidence}
\label{ss:retrieving_evidence}
For our retrieval-augmented approaches,
we experiment with five LLMs (proprietary and open-weight)
with various sizes.
Specifically, we report results with GPT-4o, Llama 3 70B, and Mistral 7B (Table \ref{t:experiment_results}) and Llama 3 8B and Qwen2 7B (Appendix~\ref{s:appendixb}).
%The prompts in Appendix~\ref{ss:appendixa} are reused with minimum modifications. 
Appendix~\ref{ss:appendixg} reports our experimental settings including hyperparameters.

Similar to the fact verification experiments,
we experiment with gold evidence to establish an (unrealistic) upper bound and two variants of retrieved evidence: retrieved using the question or statement.
Table \ref{t:experiment_results} reports results considering 10 sentences as evidence; 
Appendix~\ref{s:appendixb} provides more results considering different amounts of evidence.

LLMs benefit from retrieved evidence to identify questions with false assumptions.
This is true across all datasets and LLMs---comparing to baselines without retrieved evidence, most improvements are statistically significant 
(McNemar’s test~\cite{mcnemar1947note}, p<0.05).
In fact, LLMs incorporating retrieved evidence yield the state-of-the-art results on two datasets 
((QA)$^2$: 76.51, CREPE: 69.24). 
The previous work on FalseQA trains an LLM~\cite[MACAW-11B]{Tafjord2021Macaw} and only yields a marginal improvement 
compared to ours (86.00 vs. 83.91). 
However, we show that such a supervised approach is not transferable to other datasets 
(Appendix~\ref{s:appendixb}, Table~\ref{t:supervised_experiments}). 

Prompting with the question is mostly better than the statement.
This is due to the fact that transforming questions into statements loses information about the underlying assumptions by the author of the question.
GPT-4o outperforms the o1 model and other small size models. Bold indicates the best results for each model.

\subsection{Results Generating and Validating Atomic Assumptions}
Our approach to generate and validate atomic assumptions
is interpretable.
In addition,
it outperforms the best non-interpretable approach
(i.e., identifying with question and retrieved evidence) in most cases. 
It yields state-of-the-art results on all three datasets, some even without retrieved evidence (CREPE: 69.52).
This demonstrates that models reduce hallucinations when exposed to implicit false atomic assumptions in the question.
Importantly, Llama 3 70B gains significantly more improvements and yields comparable results to GPT-4o on two datasets (CREPE and FalseQA), showing that this approach is more beneficial for a smaller model.
Incorporating evidence is always beneficial but with a smaller margin. The trend is consistent across all models.

\begin{table*}
  \centering
  \small
  \begin{tabular}{l r l r  r}
%FP: predict valid but false
%FN: predict false but valid
\toprule

% \{5}{c}{}\\
Error Type & Dataset  & Example &  FP (\%) & FN (\%)\\
\midrule
\multirow{2}{*}{Irrelevant Evidence}  & \multirow{2}{*}{All}  & How to open the door in the house? 
 &  \multirow{2}{*}{25} & \multirow{2}{*}{23}\\
                &     & Evidence: An open house is an event held by landlords or [\ldots] \\
                \addlinespace

Relevant Evidence    \\
~~~~Wrong Label  & C, F  & Name two outdoor activities that can play indoors?  & 7  & 15 \\  \addlinespace
~~~~Ambiguous    & Q, C   & When did the Beatles get married?  &  4 & 5  \\ \addlinespace
~~~~Commonsense  & F  & How does a tenant rent the house to the owner?  & 4  & 12  \\      \addlinespace                   
~~~~Domain Knowledge & Q  & What episode does Aiden come back in Just Like That?  & 10 & 10  \\\addlinespace   
~~~~All Other	 & All & When does Korea get a new president?  & 50	& 35 \\

% ~~~~Typo in Question & Q  & When did Columbia became a country?  & 3 & 0  \\          

\bottomrule

\end{tabular}
  \caption{
  The most common error types made by our best approach 
  (Table \ref{t:experiment_results})
  in (QA)$^2$ (Q), CREPE (C), and FalseQA (F). 
  False Positive (FP) indicate the percentages of instances not having false assumptions but predicted as having false assumptions.
  False Negative (FN) indicate the opposite.
  %Percentages are the average in all corpora under FP and FN respectively (150 each).}
  }
  \label{t:error_analysis}
\end{table*}

\begin{table}
  \centering
  \small
  \setlength{\tabcolsep}{.04in}

\begin{tabular}{l rrr}

\toprule
 & (QA)$^2$ & CREPE & FalseQA \\
& (Acc)    & (F1)  & (Acc)   \\ \midrule

% Best Prev. Work & 64.00  & 0.6700  & 86.00  \\ \midrule

\multicolumn{4}{l}{Gen. \& Val. Atomic Assumptions with Llama 3 70B} \\
\addlinespace
~~~w/o Evidence (baseline) & 64.25 & 65.52 & 71.24 \\
~~~with Retrieved Evidence \\
~~~~~~Gold (upper bound)      & 76.01 & 72.71 & n/a   \\ 
~~~~~~using the Question      & \textbf{66.97} & \textbf{68.91} & 79.10\\ 
~~~~~~using the Statement     & 64.71 & 68.42 & \textbf{79.51}\\ \midrule

\multicolumn{4}{l}{Gen. \& Val. Atomic Assumptions with Mistral 7B} \\
\addlinespace
~~~w/o Evidence (baseline)    & 51.42 & \textbf{32.02} & 54.03 \\
~~~with Retrieved Evidence \\
~~~~~~Gold (upper bound)      & 61.58 & 32.96 & n/a   \\ 
~~~~~~using the Question      & \textbf{52.61} & 31.76 & \textbf{57.80}\\ 
~~~~~~using the Statement     & 52.46 & 31.92 & 55.41\\ 
\bottomrule
\end{tabular}
  \caption{
  Experimental results generating and validating atomic assumptions using Llama 3 70B and Mistral 7B.
  Llama obtains comparable results to generating atomic assumptions with GPT-4o (Table \ref{t:experiment_results}).
  %Percentages are the average in all corpora under FP and FN respectively (150 each).}
  }
  \label{t:small_model_ablation}
\end{table}

\paragraph{Generating Atomic Assumptions with Smaller LLMs}
We have demonstrated that generating (with GPT-4o)
and validating (with GPT-4o, Llama 3 70B, and Mistral 7B)
atomic assumptions yields state-of-the-art results.
However, they rely on GPT-4o to generate atomic assumptions. 
% a close-source, state-of-the-art, very large LLM to generate atomic assumptions.
%one could argue that our approach requires the state-of-the-art model (i.e., GPT-4) to generate the atomic assumptions. 

We investigate whether smaller models, Llama 3 70B and Mistral 7B, 
can both generate and validate atomic assumptions.
The generated atomic assumptions are manually verified, 
with detailed results in Appendix~\ref{ss:appendixf}.
While Llama 3 70B generates fewer atomic assumptions than GPT-4o (averaging 3.4 per question), 
Mistral 7B performs worse still (3.2 per question), reflecting its limited ability.
Table~\ref{t:small_model_ablation} shows validation results using these smaller models.
Despite generating lower-quality atomic assumptions,
Llama 3 70B achieves competitive validation performance compared to Table~\ref{t:experiment_results},
demonstrating that less perfect atomic assumptions help identify questions with false assumptions.
However, Mistral 7B shows significant performance degradation in the end-to-end pipeline,
indicating that substantially smaller models struggle with both generation and validation tasks.

\subsection{Computational Cost Analysis}
While our approach requires additional computational overhead for evidence retrieval and atomic assumption generation/validation, it remains more cost-effective than alternative methods such as prompting the o1 model.
We measure computational costs using inference tokens (input and output), with results averaged across all datasets.

Evidence retrieval incurs a one-time cost that depends on the retrieval system.
Our T5-based retriever (Appendix~\ref{ss:appendixe}) requires minimal resources.
On the other hand, 4-shot prompting without evidence consumes 151 inference tokens per question, while incorporating top-10 evidence adds 304 tokens per question.
In contrast, the o1 model requires 568 additional tokens per question due to its extensive chain-of-thought reasoning.
Moreover, o1 inference is 5 times more expensive than GPT-4o and has significantly longer query times.

Generating and validating atomic assumptions require 51 additional tokens each (102 total) and 4.7 extra queries per question.
Despite this overhead, our approach delivers superior performance with interpretable outputs while maintaining lower costs than competing methods.

\subsection{Error Analysis}
We define a False Positive (FP) as a question only having valid assumptions but predicted as having false assumptions. Similarly, we define a False Negative (FN) as a question having false assumptions but predicted as only having valid assumptions.
From the errors made by our best approach~(Table~\ref{t:experiment_results}, question and evidence retrieved with statement),
we observe that
(QA)$^2$ has a similar rate of FP (52.5\%) and FN (47.5\%) but CREPE has more FP (61.1\%) than FN (38.9\%), consistent with their label distributions 
(Table~\ref{t:dataset}).
Surprisingly, FalseQA results in significantly more FP (87.5\%) than FN (12.5\%) despite having a balanced distribution. 
We hypothesize this is due to the fact that FalseQA revises questions with false assumptions (e.g., What is the length of the air?) 
to create valid assumptions 
(e.g., What is the length of the arm?), resulting in unnatural questions.

We also conduct an error analysis to identify the most common error types by the best model. 
We analyze 50 FP and 50 FN errors from each benchmark (300 total).
Table~\ref{t:error_analysis} presents the error types.
First, 24\% of errors are due to failing to retrieve relevant evidence.
For example, the evidence retrieved for ``How to open the door in the house?'' includes the keyword ``open house'' but it is irrelevant (opening a door vs. real estate).
Wrong annotation labels account for 7\% FP and 15\% FN in CREPE and FalseQA. This issue has also been reported by~\citet{yu-etal-2023-crepe}.
(QA)$^2$ and CREPE include ambiguous questions that do not necessarily have false assumptions since they expect multiple valid answers.
8\% of questions from FalseQA require simple commonsense knowledge to identify false assumptions, however, the model fails to do so even with the help of retrieved evidence. 
Finally, 10\% errors in (QA)$^2$ require domain knowledge
(e.g., understanding TV show plots).

\section{Validating Atomic Assumptions Provides Interpretations}
\label{s:atomic_fact}

\noindent
\textbf{Are LLMs Capable of Generating Interpretations?} 
%\todo{new section: directly generate the interpretation}
Before presenting our approach that validates atomic assumptions,  
we investigate if LLMs have the ability to directly generate an interpretation for a question with false assumptions. 
Specifically, we prompt two LLMs, GPT-4o and Llama 3 70B, to generate interpretations.
We do not include smaller models as they have already demonstrated limited ability in previous tasks.
We randomly choose 100 questions with false assumptions per benchmark that are successfully identified by our approach 
(GPT-4o incorporating evidence retrieved with question, Table~\ref{t:experiment_results}).
Note that we also provide in the prompt that the question has false assumptions.
The generated interpretation is evaluated manually by checking whether it pinpoints the false assumptions. 
Appendix~\ref{ss:appendix_evaluation} details the evaluation process.
GPT-4o yields an accuracy of 0.86, 0.67, and 0.93 on (QA)$^2$, CREPE, and FalseQA, respectively, 
and Llama 3 70B yields an accuracy of 0.81, 0.66, and 0.84, respectively, 
showing that LLMs still hallucinate when generating interpretations even with knowing the question has false assumptions.
Besides performance, this approach requires more computational costs than our approach,
with an average of 121 tokens per question.
Appendix~\ref{ss:detail_interpretation} reports the prompts and exemplifies errors by the two models.

We now evaluate our approach. Does validating atomic assumptions provide interpretations? 
We reuse the same questions from the previous study
and manually annotate the interpretation of the atomic assumptions:
which atomic assumptions are true and false.
Appendix~\ref{ss:appendix_annotation} reports details including inter-annotator agreement.

The benchmark contains 1,006 atomic assumptions,
of which 534 are false and 472 are true.
% We evaluate our approach using weighted F1 Score. 
The results show strong performance with F1 scores of 0.86 for (QA)$^2$, 0.88 for CREPE, and 0.87 for FalseQA,
demonstrating that it successfully provides interpretations by pinpointing specific false assumptions.
Appendix~\ref{ss:appendix_add_results} reports detailed results including Precision and Recall for each label.

\section{Conclusions}
Identifying and answering questions with false assumptions is a challenging task for state-of-the-art LLMs.
The main issue is that LLMs are overconfident and hallucinate answers to these kinds of questions.
Additionally, fact verification cannot solve this problem as false assumptions often do not challenge factual information.

We introduce an approach that leverages evidence retrieval to mitigate hallucinations. 
Experimental results show it is beneficial for this task.
Crucially, the benchmarks span several
domains (Reddit, search queries, etc.)
and
procedures to introduce false assumptions (genuine user-generated, crowdsourcing, etc.). 
Validating atomic assumptions derived from a question yields state-of-the-art results on all three datasets.
Most importantly, it provides human-readable interpretations of the false assumptions
beyond simply determining whether a question has a false assumption.
These interpretations pinpoint the specific assumption that is false and the many assumptions that are true in a question.

\section*{Limitations}

Our experimental methodology relies primarily on LLM prompting, 
which inherently limits reproducibility due to the nature of large language models.
To mitigate this concern, we provide comprehensive implementation details, including exact prompts and experimental settings.
Furthermore, we validate our findings across five LLMs of varying sizes, encompassing both proprietary and open-weight models, 
to ensure the generalizability of our results.

The performance of our approach depends on the underlying retrieval system, 
which could be viewed as a limitation.
However, we consider this modularity advantageous, 
as the retrieval component can be seamlessly replaced or upgraded as better systems become available.

Our methodology incurs additional computational overhead compared to baseline approaches.
Specifically, retrieving evidence requires approximately 2 times more tokens than question-only identification, 
while generating and validating atomic assumptions demands 33\% more tokens and 3.7 times more queries.
Despite this overhead, our approach remains substantially more efficient than alternative methods such as prompting o1 models or direct interpretation generation, which require significantly higher computational costs.

\section*{Ethics Statement}

\textbf{Data Sources and Collection}
We collect the dataset ((QA)$^2$, CREPE, FalseQA, BoolQ, and FEVER) via the link provided by the authors.

Data from (QA)$^2$~\cite{kim-etal-2023-qa} is used under Apache-2.0 License;
CREPE~\cite{yu-etal-2023-crepe} under BSD License; BoolQ~\cite{clark-etal-2019-boolq} under CC BY-SA 3.0 License; 
and FEVER~\cite{thorne-etal-2018-fever} under Apache-2.0 License.
FalseQA~\cite{hu-etal-2023-wont} does not specify their license.

\section*{Acknowledgments}

We would like to thank the Chameleon platform~\cite{keahey2020lessons} for providing computational resources. 
We also thank the reviewers for their insightful comments.

The Microsoft Accelerating Foundation Models Research Program and the OpenAI Researcher Access Program provided credits
to conduct this research.
This material is based upon work supported
by the National Science Foundation under Grant
No. 2310334. Any opinions, findings, and conclusions or recommendations expressed in this material are those of the authors and do not necessarily
reflect the views of the NSF.
\bibliography{custom}

\appendix

\section{Additional Details to Identify Questions with False Assumptions}
\label{s:appendixa}

\subsection{Prompts to Transform Questions into Statements}
\label{ss:appendixb}
Figure~\ref{f:prompts2} reports the prompts to transform questions into statements.

\begin{figure}[!h]
    \small
    \begin{framed}
    \texttt{You will be provided with a question. Your task is to transform the question into a statement and keep its original meaning.} \\
    
    \texttt{Question: How do hashing functions avoid collisions?}\\
    \texttt{Statement: Hashing functions can avoid collisions.}\\
    
    \texttt{Question: Who is the only Indian to win the Oscar for music?}\\
    \texttt{Statement: Only one Indian has won the Oscar for music.}\\
    
    \texttt{Question: Why have our bodies arrived at 98.6F as the ``normal'' body temperature?}\\
    \texttt{Statement: 98.6F is the ``normal'' body temperature.}\\
    
    \texttt{Question: What kind of meat can be made into soybean milk?}\\
    \texttt{Statement: Soybean milk can be made from meat.}\\
    
    \texttt{Question: \{question\}}\\
    \texttt{Statement: \{\}}
    \end{framed}
    \caption{4-shot prompts to transform questions into statements.}
    \label{f:prompts2}
\end{figure}

\subsection{Prompts to Identify False Assumptions}
\label{ss:appendixa}

Figure~\ref{f:prompts1} reports the complete version of our prompts to identify false assumptions in questions.
The few-shot examples are sampled from the training splits.
\begin{figure}[!h]
    \small
    \begin{framed}
    \texttt{You are a helpful assistant that helps identify false assumptions. 
    Output Yes if the \{question | statement\} has false assumptions; otherwise, output No.}\\\\
    \texttt{Input: \{How do betta fish survive without oxygen? | Betta fish can survive without oxygen.\}\\}
    \texttt{Question: Does the input contain any false assumptions?\\}
    \texttt{Answer: Yes\\}

    \texttt{Input: \{Who is the Duke of Oxford? | There exists the Duke of Oxford.\}\\}
    \texttt{Question: Does the input contain any false assumptions?\\}
    \texttt{Answer: No\\}
    
    \texttt{Input: \{Where does the Flint River in Georgia start and end? | The Flint River in Georgia start and end in someplace.\}\\} 
    \texttt{Question: Does the input contain any false assumptions?\\}
    \texttt{Answer: No\\}

    \texttt{Input: \{Who is the movie Jersey based on? | The movie Jersey is based on someone.\}  \\}
    \texttt{Question: Does the input contain any false assumptions?\\}
    \texttt{Answer: Yes\\}

    \texttt{Input: \{question | statement\}}\\
    \texttt{Question: Does the input contain any false assumptions?}\\
    \texttt{Answer: \{\}}
    \end{framed}
    \caption{The complete version of prompts to identify false assumptions in questions (or statements).}
      \label{f:prompts1}
    \end{figure}

\subsection{Prompts to Generate Evidence}
\label{ss:appendixc}
Figure~\ref{f:prompts3} reports the prompts to  generate relevant knowledge based on the question.
We reuse the same prompts provided by~\citet{liu-etal-2022-generated} with minimal modifications.

\begin{figure}[!h]
    \small
    \begin{framed}
    \texttt{Generate some knowledge about the input.}\\\\
    \texttt{Input: Greece is larger than Mexico.}\\
    \texttt{Knowledge: Greece is approximately 131,957 sq km, while Mexico is approximately 1,964,375 sq km, making Mexico 1,389\% larger than Greece.}\\
   
    \texttt{Input: A fish is capable of thinking.}\\
    \texttt{Knowledge: Fish are more intelligent than they appear. In many areas, such as memory, 
    their cognitive powers match or exceed those of `higher' vertebrates including non-human primates.}\\
    
    \texttt{Input: A common effect of smoking lots of cigarettes in one's lifetime is a higher than normal chance of getting lung cancer.}\\
    \texttt{Knowledge: Those who consistently averaged less than one cigarette per day over their lifetime had nine times the risk of dying from lung cancer than never smokers. 
    Among people who smoked between one and 10 cigarettes per day, the risk of dying from lung cancer was nearly 12 times higher than that of never smokers.}\\
    
    \texttt{Input: A rock is the same size as a pebble.}\\
    \texttt{Knowledge: A pebble is a clast of rock with a particle size of 4 to 64 millimeters based on the Udden-Wentworth scale of sedimentology. 
    Pebbles are generally considered larger than granules (2 to 4 millimeters diameter) and smaller than cobbles (64 to 256 millimeters diameter).}\\

    \texttt{Input: \{question\}}\\
    \texttt{Knowledge: \{\}}
    \end{framed}
    \caption{4-shot prompts to generate relevant knowledge based on the question.}
    \label{f:prompts3}
\end{figure}

\subsection{Evidence Retrieval Details}
\label{ss:appendixe}

% \noindent
% \textbf{Retrieving Documents}
% We start by retrieving documents relevant to the question (or statement).
% CREPE includes relevant passages from Wikipedia retrieved with C-REALM~\cite{krishna-etal-2021-hurdles} and we adopt them.
% For (QA)$^2$ and FalseQA, we query the Google Search Engine API,
% limiting the results to Wikipedia. 
% We retain the top three articles.

% \noindent
% \textbf{Retrieving Sentences}
% After obtaining relevant Wikipedia articles,
% we consider all sentences as candidate evidence.
% %identify sentence boundaries. 
% %Each sentence is taken as a candidate evidence to the question.
% We use INSTRUCTOR~\cite{su-etal-2023-one}, a state-of-the-art text embedding model, to retrieve the top $k$ sentences most relevant to the question (or statement) from the Wikipedia articles.
% We take $k$ as a parameter to be tuned (max: 10).
% Whether the retrieved evidence is useful for identifying false assumptions is an empirical question that can be answered based on experimental results.
% Table~\ref{t:dataset_example} presents questions and retrieved evidence.
% For example, the evidence for ``When did they stop using lead in pencils'' 
% includes ``[\ldots] lead has not been used for writing'', which contradicts the false assumption in the question (i.e., ``People once used lead in pencils'').

\paragraph{Experimental Setting}
To retrieve relevant document, we query the question (or the statement) using Google Search Engine API,\footnote{\url{https://developers.google.com/custom-search}}
and limit the results to Wikipedia. The retrieved Wikipedia documents are further parsed to only retain the main content.
We use the INSTRUCTOR model~\cite{su-etal-2023-one}  to rank the sentences based on the similarity to the question (or statement). 
The input to the model include: (1) sentences from Wikipedia pages retrieved from the last step (Section~\ref{ss:retrieving}),
and (2) the question or statement we want to retrieve with.
Figure~\ref{f:instructions} reports the instructions for the INSTRUCTOR model to 
rank the candidate sentences, based on the similarity to the question or statement.

\begin{figure}
    \small
    \begin{framed}
    \texttt{Represent the \{question | statement\} for retrieving supporting evidence: \{question\}}\\\\
    \texttt{Represent the evidence for retrieval: \{passages from Wikipedia\}}

    \end{framed}
    \caption{Instructions to identify questions or statements.}
      \label{f:instructions}
\end{figure}

\subsection{Prompts for Identifying False Assumptions with Retrieved Evidence}
\label{ss:appendixd}
Figure~\ref{f:prompts4} reports the 4-shot prompts to identify false assumptions with retrieved evidence.
Note that due to resource limitations, we are only able to add one evidence per example for the 4-shot examples.

\begin{figure}[h!]
    \small
    \begin{framed}
    \texttt{You are a helpful assistant that helps identify false assumptions in \{question | statement\}. 
    Use the information from the evidence to help you identify the false assumption.
    Output Yes if the \{question | statement\} has false assumptions; otherwise, output No.}\\
    \texttt{Input: Why can't we vote online? It seems ridiculous we have to drive to do such a simple and important task.}\\
    \texttt{Evidence: Many countries have looked into Internet voting as a possible solution for low voter turnout. Some countries like France and Switzerland use Internet voting.}\\
    \texttt{Question: Considering the external knowledge from the Evidence, does the input contain any false assumptions?\\}
    \texttt{Answer: Yes\\}

    \texttt{Input: Why are trees susceptible to lightning strikes?}\\
    \texttt{Evidence: Direct strike casualties could be much higher than reported numbers. Trees are frequent conductors of lightning to the ground.}
    \texttt{Question: Considering the external knowledge from the Evidence, does the input contain any false assumptions?\\}
    \texttt{Answer: No\\}

    \texttt{Input: Why does a bad throat often turn to common cold?}\\
    \texttt{Evidence: The distinction between viral upper respiratory tract infections is loosely based on the location of symptoms with the common cold affecting primarily the nose, pharyngitis (the throat), and bronchitis (the lungs).}\\
    \texttt{Question: Considering the external knowledge from the Evidence, does the input contain any false assumptions?\\}
    \texttt{Answer: Yes\\}

    \texttt{Input: Why does clear plastic turn opaque and white when bent?}\\
    \texttt{Evidence: Stress-whitening is where a white line appears along a bend or curve when a material is stressed by bending or punching operations.}\\
    \texttt{Question: Considering the external knowledge from the Evidence, does the input contain any false assumptions?\\}
    \texttt{Answer: Yes\\}  

    \texttt{Input: \{question | statement \}}\\
    \texttt{Evidence: \{\}}\\
    \texttt{Question: Considering the external knowledge from the Evidence, does the input contain any false assumptions?\\}
    \texttt{Answer: \{\}}
    \end{framed}
    \caption{4-shot prompts to identify false assumptions with retrieved evidence.}
    \label{f:prompts4}
\end{figure}

\subsection{Generating and Validating Atomic Assumptions}
\label{ss:appendixf}
\noindent\paragraph{GPT-4o Model}
Figure~\ref{f:prompts5} reports the prompts used to extract atomic assumptions from the questions.
Note that we reuse the prompts in Figure~\ref{f:prompts1} to identify false assumptions in atomic assumptions.
Similar to the transformation of questions into statements,
we evaluate the generated atomic assumptions by manually checking 200 random questions from each benchmark (600 total).
The precision of generated atomic assumptions is near perfect for all three datasets ((QA)$^2$: 0.98, CREPE: 0.95, and FalseQA: 0.95). 
We generate 2,410, 14,701 and 6,140 atomic assumptions for (QA)$^2$ (570 questions), CREPE (3,004 questions), and FalseQA (1,374 questions) respectively,
resulting in 4.7 atomic assumptions per question.

\noindent\paragraph{Llama and Mistral Model}
We use the same prompt as in Figure~\ref{f:prompts5} to generate atomic assumptions with Llama 3 70B and Mistral 7B.
Llama 3 70B shows worse yet competitive ability
compared to GPT-4o ((QA)$^2$: 0.90, CREPE: 0.87, and FalseQA: 0.90), 
and Mistral 7B yields worse results (0.82, 0.83, and 0.75 respectively).
They yield on average fewer atomic assumptions per question (Llama 3: 3.4, and Mistral: 3.2).
\begin{figure}
    \small
    \begin{framed}
    \texttt{ You are a helpful assistant. Help me understand the question by extracting both explicit and implicit atomic assumptions. 
    You must notice that considering the intention of the question asker is helpful for extracting a hidden assumption. 
    Output every atomic assumption in a complete sentence.}\\

    \texttt{Question: When did the great depression begin before world war 1?}\\
    \texttt{Let us think step by step,the atomic assumptions are:}\\
    \texttt{(1) There was a period called the Great Depression.}\\
    \texttt{(2) There was a conflict called World War 1.}\\
    \texttt{(3) The Great Depression began before World War 1.}\\
    
    \texttt{Question: How do betta fish survive without oxygen?}\\
    \texttt{Let us think step by step,the atomic assumptions are:}\\
    \texttt{(1) There is a type of fish called betta fish.}\\
    \texttt{(2) Fish can survive without oxygen.}\\
    
    \texttt{Question: Why is card counting against the rules at casinos?}\\
    \texttt{Let us think step by step,the atomic assumptions are:}\\
    \texttt{(1) Card counting is a strategy used at casinos.}\\
    \texttt{(2) Casinos can have rules against certain behaviors.}\\
    \texttt{(3) Card counting is not allowed in some places.}\\
        
    \texttt{Question: How does the chest cavity close up after heart surgery is performed?.}\\
    \texttt{Let us think step by step,the atomic assumptions are:}\\
    \texttt{(1) The chest cavity can be opened and then closed up.}\\
    \texttt{(2) Heart surgery requires opening of the chest cavity.}\\
    \texttt{(3) The close of the chest cavity happens after heart surgery.}\\

    \texttt{Question: \{question\}}\\
    \texttt{Let us think step by step,the atomic assumptions are:}
   
    \end{framed}
    \caption{4-shot Chain-of-Thought prompts to extract atomic assumptions from the questions.}
    \label{f:prompts5}
\end{figure}

\subsection{Experimental Settings}
\label{ss:appendixg}

\noindent
\paragraph{Fact Verification}
We use MiniCheck~\cite{tang-etal-2024-minicheck}, 
the state-of-the-art fact verification system according to LLM-AggreFact Leaderboard.\footnote{\url{https://llm-aggrefact.github.io/}}
Specifically, we access the model (Llama-3.1-Bespoke-MiniCheck-7B) via the API hosted by Bespoke Labs.\footnote{\url{https://playground.bespokelabs.ai/}}
The returned \texttt{support\_prob} score is mapped to our labels (valid or false assumption) using a threshold of \texttt{0.5}.

\noindent
\paragraph{Supervised Approach}
We train an off-the-shelf RoBERTa-large model~\cite{liu2019roberta} (355M parameters)
from Hugging Face~\cite{wolf-etal-2020-transformers}. 
The experiments are conducted on a single NVIDIA Tesla V100 (32GB) GPU. 
A single experiment takes approximately 1 hour for training, 
but the time may vary depending on the training dataset size. 
Table~\ref{t:roberta_hyperparameters} reports the hyperparameters used in the experiments.
\begin{table}
    \centering
    \small

\begin{tabular}{  p{1.4in} r}

\toprule
& RoBERTa-large model \\
 \midrule

Maximum Epochs & 20\\
Batch Size & 32\\
Optimizer & AdamW \\
Learning Rate  &1e-5 \\
Weight Decay &  0.01 \\

    \bottomrule
    
    \end{tabular}
    \caption{
     Hyperparameters used in our supervised approach with RoBERTa-large model. 
     We adopt AdamW~\cite{loshchilov2018decoupled} as the optimizer.
    }
    \label{t:roberta_hyperparameters}
  \end{table}

\noindent
\paragraph{Prompting LLMs}
We prompt GPT-4o (2024-08-07) for two experiments: (1) transforming questions into statements (Section~\ref{ss:baseline}) 
and (2) generating atomic assumptions from the questions (Section~\ref{ss:generating}).
Another five LLMs (GPT-4o (2024-08-07), Mistral-7B-Instruct-v0.3, Qwen2-7B-Instruct, Llama-3-8B-Instruct, and Llama-3-70B-Instruct) 
are used for: (1) identifying questions with false assumptions with and without evidence, and (2) validating atomic assumptions (both in Table~\ref{t:experiment_results}).
GPT-4o (2024-08-07) is further used to identifying false assumptions with generated evidence (Table~\ref{t:baseline}).
We access GPT-4o API via Microsoft Azure AI.\footnote{\url{https://azure.microsoft.com/solutions/ai}}
We host the other four LLMs, Mistral-7B-Instruct-v0.3, Qwen2-7B-Instruct, Llama-3-8B-Instruct, and Llama-3-70B-Instruct via deepinfra.\footnote{\url{https://deepinfra.com}}
We set the \texttt{temperature} as 0.1, \texttt{top\_p} as 0.1, and \texttt{frequency\_penalty} as 0 for all experiments.
The maximum generation length is set to 4 or 512 tokens depending on the tasks (i.e., identifying or answering questions with false assumptions). 

\section{Additional Results Identifying Questions with False Assumptions}
\label{s:appendixb}

For our supervised baseline experiments, we only report the best results on each corpus in Table~\ref{t:baseline}.
The complete results are reported in Table~\ref{t:supervised_experiments}.

\begin{table}
  \centering
  \small
  \setlength{\tabcolsep}{.04in}

\begin{tabular}{l rrr}

\toprule
 & (QA)$^2$ & CREPE & FalseQA \\
& (Acc)    & (F1)  & (Acc)   \\ \midrule

Best Prev. Work & 0.64  & 0.67  & 0.86  \\ \midrule

RoBERTa trained with \\
~~~CREPE           & 0.50  & 0.60  & 0.50  \\
~~~FalseQA         & 0.56  & 0.55  & 0.71  \\
~~~CREPE + FalseQA & 0.49  & 0.62  & 0.52  \\
~~~~~~+ BoolQ      & 0.52  & 0.52  & 0.72  \\
~~~~~~+ FEVER      & 0.55  & 0.58  & 0.71  \\
~~~All             & 0.52  & 0.59  & 0.73  \\
\bottomrule

\end{tabular}
  \caption{
    The complete results to identify false assumptions from our supervised baseline.
    We train a RoBERTa-large model with several related datasets including BoolQ and FEVER.
  }
  \label{t:supervised_experiments}
\end{table}

Since Mistral 7B yields the best performance among similar size LLMs (Llama 3 8B and Qwen2 7B), we only report the results with Mistral 7B in Table \ref{t:experiment_results} in the main paper.
Table~\ref{t:additional_model_results} contains extra results with Llama 3 8B and Qwen2 7B.
The conclusion is the same across all models---it is beneficial to incorporate extra evidence.

Our evidence retrieval experiments take into account the 10 most relevant sentences per question.
We take the number of sentences in the evidence as a hyperparameter to be tuned.
Due to space limitations, Table \ref{t:experiment_results} only reports the results with top 10 sentences.
Table~\ref{t:additional_results_topk} reports additional results taking into account other numbers of sentences as evidence: top 1, 5, and 10.

\begin{table*}
  \centering
  \small
\begin{tabular}{c@{\hspace{0.1cm}}c@{\hspace{0.2cm}}c}

    %subtable 0 
    \begin{tabular}{l}
        \toprule
        \\
        \\  \midrule
    Best from Baselines   \\ \midrule
    Identifying with Question  \\
    ~~~w/o Evidence (baseline) \\
    ~~~with Retrieved Evidence \\
    ~~~~~~Gold (upper bound)                     \\
    ~~~~~~using the Question      \\
    ~~~~~~using the Statement     \\ \addlinespace
    Identifying with Statement \\
    ~~~w/o Evidence (baseline)  \\
    ~~~with Retrieved Evidence \\
    ~~~~~~Gold (upper bound)                  \\
    ~~~~~~using the Question     \\
    ~~~~~~using the Statement    \\ \midrule
    Gen. \& Val. Atomic Assumptions \\
    ~~~w/o Evidence  \\
    ~~~with Retrieved Evidence \\
    ~~~~~~Gold (upper bound)     \\
    ~~~~~~using the Question   \\
    ~~~~~~using the Statement \\
    \bottomrule
    \end{tabular}
    
    &

    % subtable1
    \begin{tabular}{ rrr}
    
    \toprule
    (QA)$^2$ & CREPE & FalseQA \\
    Acc) & (F1) & (Acc) \\ \midrule
    
    0.75    & 0.67  & 0.81 \\ \midrule
    %(Table~\ref{t:baseline})
    \\
     0.48  & 0.45  & 0.52  \\
    \\
    0.61  & 0.57  &  n/a \\
    0.52  & 0.55  & 0.56 \\
    0.53  & 0.55  & 0.57 \\ \addlinespace
    
    \\
    0.48  & 0.49  & 0.56 \\
     \\
    0.63  & 0.56  &  n/a \\
    0.52  & 0.54  & 0.55 \\
    0.54  & 0.52  & 0.55 \\ \midrule
    
    \\
    0.51  & 0.34  & 0.63  \\
    \\
    0.68  & 0.55  &  n/a  \\
    0.59  & 0.49  & 0.66  \\
    0.61  & 0.47  & 0.63  \\
    
        \bottomrule
        
        \end{tabular}
    
    & 
    
    %subtable 2
    
    \begin{tabular}{ rrr}
    
        \toprule
        (QA)$^2$ & CREPE & FalseQA \\
        (Acc) & (F1) & (Acc) \\ \midrule
        
         0.75    & 0.67  & 0.81  \\ \midrule
        %(Table~\ref{t:baseline})
         \\
         0.53  & 0.45  & 0.50  \\
         \\
         0.54  & 0.48  &  n/a \\
         0.50  & 0.56  & 0.51 \\
         0.51  & 0.56  & 0.51 \\ \addlinespace
        
         \\
         0.52  & 0.46  & 0.53 \\
          \\
         0.62  & 0.52  &  n/a \\
         0.51  & 0.56  & 0.51 \\
         0.51  & 0.55  & 0.51 \\ \midrule
        
         \\
         0.49  & 0.46  & 0.55  \\
         \\
         0.66  & 0.60  &  n/a  \\
         0.54  & 0.53  & 0.61  \\
         0.51  & 0.51  & 0.61  \\
        
            \bottomrule
            
            \end{tabular}\\\\

          &  (a) Llama 3 8B & (b) Qwen2 7B \\
    \end{tabular}
  \caption{
    Experimental results to identify false assumptions with Llama 3 8B and Qwen2 7B. For experiments incorporating evidence, we choose the top 10 evidence.
  }
  \label{t:additional_model_results}
\end{table*}

\begin{table}[h!]
    \centering
    \small
    \setlength{\tabcolsep}{.051 in}

\begin{tabular}{ l rrr}

\toprule
& (QA)$^2$ & CREPE & FalseQA \\
& (Acc) & (F1) & (Acc) \\ \midrule

Identifying with Question \\
~~~Retrieved w/ Question   \\  
~~~~~~top 1 sentence         & 0.70  & 0.56  & 0.66 \\
~~~~~~top 5 sentence         & 0.72  & 0.67  & 0.75 \\
~~~~~~top 10 sentence        & 0.76  & 0.69  & 0.84 \\

~~~Retrieved w/ Statement  \\   
~~~~~~top 1 sentence         & 0.70  & 0.57  & 0.67\\
~~~~~~top 5 sentence         & 0.73  & 0.64  & 0.72\\
~~~~~~top 10 sentence        & 0.76  & 0.68  & 0.83 \\ 
\midrule
Identifying with Statement \\
~~~Retrieved w/ Question    \\
~~~~~~top 1 sentence         & 0.74  & 0.51  & 0.74\\
~~~~~~top 5 sentence         & 0.74  & 0.59  & 0.76\\
~~~~~~top 10 sentence        & 0.77  & 0.64  & 0.79 \\
~~~Retrieved w/ Statement   \\
~~~~~~top 1 sentence         & 0.74   & 0.52  & 0.73 \\
~~~~~~top 5 sentence         & 0.75   & 0.54  & 0.75 \\
~~~~~~top 10 sentence        & 0.76  & 0.63  & 0.79 \\ 
\midrule

Validating Atomic Assumptions \\
~~~retrieved w/ Question  \\
~~~~~~top 1 sentence         & 0.68  & 0.48  & 0.65\\
~~~~~~top 5 sentence         & 0.72  & 0.60  & 0.77\\
~~~~~~top 10 sentence        & 0.74  & 0.70  & 0.86  \\
~~~retrieved w/ Statement \\
~~~~~~top 1 sentence         & 0.69  & 0.49  & 0.64 \\
~~~~~~top 5 sentence         & 0.73  & 0.61  & 0.79 \\
~~~~~~top 10 sentence        & 0.73  & 0.68  & 0.85  \\

    \bottomrule
    
    \end{tabular}
    \caption{
      Experimental results with GPT-4o to identify false assumptions with different amounts of retrieved evidence.
      We take the number of sentences in the evidence as a tuned parameter (1, 5, and 10). Overall, including more sentences yields better results.
  %  Experiment results on verifying questions with false assumptions. We report the same metric as the original paper
  %  for each dataset.
    }
    \label{t:additional_results_topk}
  \end{table}

%%%%%%%%%%%%%%%%%%%%%%%%%%%%%%%%%%%%%%%%%%%%%%%%  
\section{Additional Details on Generating Interpretations via Validating Atomic Assumptions}
\label{s:appendixc}

\subsection{Evaluating Generated Interpretations}
\label{ss:appendix_evaluation}
We recruit two graduate students to evaluate if the generated interpretation correctly pinpoints the false assumptions in the questions.
The gold answer from the original dataset is provided to avoid any misunderstanding of the question. 
The generated interpretation is evaluated as \textit{False} if it (1) fails to pinpoint the false assumptions, or (2) includes any additional false facts.
Accuracy is used to calculate the evaluation results.
The overall inter-annotator agreements (Cohen's $\kappa$) between two annotators are, (QA)$^2$: 0.65, CREPE: 0.68, and FalseQA: 0.73, showing substantial agreement.

\subsection{Details and Error Analysis of Generated Interpretations}
\label{ss:detail_interpretation}

\paragraph{Prompts Used to Generate Interpretations}
Figure~\ref{f:prompt_for_interpretation} provides the 4-shot prompts used to generate the interpretation to pinpoint the false assumptions in the question.

\begin{figure}
  \small
  \begin{framed}
  \texttt{You will be provided with a question that contains at least 1 false assumption. Your task is to help me understand what are the false assumptions.
  Write an explanation to pinpoint the false assumptions.}\\

  \texttt{Question: How do betta fish survive without oxygen?}\\
  \texttt{Evidence: \ldots}\\
  \texttt{Interpretation: The question has a false assumption that betta fish can survive without oxygen. In fact, they need oxygen to survive.}\\

  \texttt{Question: Why can't we vote online? }\\
  \texttt{Evidence: \ldots}\\
  \texttt{Interpretation: The question has a false assumption that people cannot vote online. However, in some countries, it is allowed to vote online.}\\
  
  \texttt{Question: \ldots }\\
  \texttt{Evidence: \ldots}\\
  \texttt{Interpretation: \ldots}\\

  \texttt{Question: \ldots }\\
  \texttt{Evidence: \ldots}\\
  \texttt{Interpretation: \ldots}\\
  
  \texttt{Question: \{question\}}\\
  \texttt{Evidence: \{evidence\}}\\
  \texttt{Interpretation: \{\}}

  \end{framed}
  \caption{4-shot prompts to directly generate the interpretation, after identified the question has false assumptions.}
    \label{f:prompt_for_interpretation}
\end{figure}

\paragraph{Error Interpretations Generated by LLMs}
Table~\ref{t:error_example_interpretation} lists three error examples that made by the two LLMs (GPT-4o and Llama 3 70B) we evaluated.
The errors are mainly two types: (1) fail to pinpoint the false assumption, or (2) include additional false facts in the claim.

\begin{table*}
  \centering
  \small
  \begin{tabular}{l@{ } p{12 cm} }

\toprule

% \{5}{c}{}\\
% Question & False Assumption & Correction & Generated Interpretation  \\
% \midrule
Question: & Can our eyes only see up to 3D, or are 4D+ objects just unable to be seen/created? \\
False Assumption: & Human eyes can see up to 3D.\\
Correction:  & Human eyes can only see 2D images. \\
Generated Interpretation: & The question has a false assumption that 4D+ objects are unable to be seen or created because our eyes can only see up to 3D. In reality, our eyes and brain are adapted to perceive the world in three dimensions (length, width, and height). [...] \\

\addlinespace

Question: & When glaciers melt during the warmer months how much gets frozen back during the cooler months? \\
False Assumption: & Glaciers freeze the ice that has melted off previously to regain size. \\
Correction: & Glaciers~do~not~refreeze~ice,~glaciers accumulate snow throughout the year, which allows them to gain size. \\
Generated Interpretation: & The question has a false assumption that all the ice lost from glaciers during the warmer months is fully regained during the cooler months. In reality, due to climate change, many glaciers are losing more ice in the warmer months than they regain in the cooler months, leading to a net loss over time.\\
\addlinespace
Question: & When did horses stop being used for glue? \\ 
False Assumption: & Horses have stopped being used for glue \\
Correction: & Animals including horses are still used for glue, although animal glue has been mostly replaced by other adhesives and plastics. Animal glue is still manufactured and used for purposes such as restoration.\\
Generated Interpretation: & The question has a false assumption that horses were ever used to make glue. In reality, horses have never been used to make glue. The common misconception likely stems from the fact that horse hooves and bones were used to make a type of glue called ``hide glue'' in the past, but not the horses themselves. \\

\bottomrule

\end{tabular}
  \caption{
  Three error examples of generated interpretations made by LLMs. We also list the correct false assumption and its correction for comparison.
  }
  \label{t:error_example_interpretation}
\end{table*}

\subsection{Benchmark Annotation Details}
\label{ss:appendix_annotation}
We create a benchmark to evaluate whether validating atomic assumptions
provides interpretations.
The benchmark annotation process evaluates atomic assumptions independently of the original questions.
Annotators are asked to verify whether the atomic assumptions are false using any available methods (e.g., online search).
We discard questions whose atomic assumptions cannot be assigned a valid label.
The annotations were conducted in-house by two graduate students.
The inter-annotator agreements (Cohen's $\kappa$) for (QA)$^2$, CREPE, and FalseQA are 0.84, 0.67, and 0.81, respectively,
indicating substantial to perfect agreement~\cite{artstein-poesio-2008-survey}.

\paragraph{Annotator Demographics}
Two graduate students, including a female and a male, volunteered to conduct the annotations.
Both of them are Asian and have research experience in Computer Science.

\subsection{Additional Results to Validate the Atomic Assumptions}
\label{ss:appendix_add_results}
Table~\ref{t:additional_results_pr} reports additional results including Precision and Recall for each label.

\begin{table*}[h!]
    \centering
    \small

\begin{tabular}{ l rr r rr r rr}

\toprule
& \multicolumn{8}{c}{(QA)$^2$} \\
\cmidrule{2-9} 
& \multicolumn{2}{c}{Valid}  
&& \multicolumn{2}{c}{False} 
&& \multicolumn{2}{c}{All}\\
\cmidrule{2-3} \cmidrule{5-6} \cmidrule{8-9} 
& P & R  && P & R && P & R\\

Validating Atomic Assumptions 
   
& 0.92 & 0.81  && 0.81 & 9.92 &&  0.87 & 0.86  \\
    \bottomrule
    
    \end{tabular}

\noindent\begin{tabular}{ l rr r rr r rr} 
    \addlinespace 
    \toprule
    & \multicolumn{8}{c}{CREPE} \\
    \cmidrule{2-9} 
    & \multicolumn{2}{c}{Valid}  
    && \multicolumn{2}{c}{False}
    && \multicolumn{2}{c}{All}\\ 
    \cmidrule{2-3} \cmidrule{5-6} \cmidrule{8-9} 
    & P & R  && P & R && P & R\\
    Validating Atomic Assumptions 
    & 0.87 & 0.88 && 0.88 & 0.88 &&0.88 & 0.88   \\
    \bottomrule
\end{tabular}

\noindent\begin{tabular}{ l rr r rr r rr}  
    \addlinespace
    \toprule
    & \multicolumn{8}{c}{FalseQA} \\
    \cmidrule{2-9} 
    & \multicolumn{2}{c}{Valid} 
     && \multicolumn{2}{c}{False}
     && \multicolumn{2}{c}{All}\\ 
    \cmidrule{2-3} \cmidrule{5-6} \cmidrule{8-9} 
    & P & R  && P & R && P & R\\
    Validating Atomic Assumptions 
    & 0.93 & 0.72 && 0.84 &0.96 &&  0.88 & 0.87\\
    \bottomrule
\end{tabular}
    \caption{
      Results to validate atomic assumptions. We report metrics including Precision (P) and Recall (R) for each label.
  %  Experiment results on verifying questions with false assumptions. We report the same metric as the original paper
  %  for each dataset.
    }
    \label{t:additional_results_pr}
  \end{table*}

\end{document}